\documentclass[table,dvipsnames,svgnames]{article}  % or whatever document class you're using

\usepackage{aaai25}  % DO NOT CHANGE THIS
\usepackage{times}  % DO NOT CHANGE THIS
\usepackage{helvet}  % DO NOT CHANGE THIS
\usepackage{courier}  % DO NOT CHANGE THIS
\usepackage[hyphens]{url}  % DO NOT CHANGE THIS
\usepackage{graphicx} % DO NOT CHANGE THIS
\usepackage{natbib}  % DO NOT CHANGE THIS AND DO NOT ADD ANY OPTIONS TO IT
\usepackage{caption} % DO NOT CHANGE THIS AND DO NOT ADD ANY OPTIONS TO IT
\usepackage{algorithm}
\usepackage{xspace} % Include the xspace package
\usepackage{graphicx}
\usepackage{multirow}
\usepackage{booktabs}
\usepackage{adjustbox} % Make sure to include this package
\usepackage{subcaption}
\usepackage{caption}
\usepackage{pdflscape} % or \usepackage{lscape}
\usepackage[table]{xcolor}
\usepackage{verbatim}
\usepackage{mathptmx} % This is Times font
\usepackage{amsmath,amssymb,amsfonts}
\usepackage[noend]{algpseudocode}
\usepackage{xcolor, soul}
\sethlcolor{yellow}
\usepackage{siunitx}

\PassOptionsToPackage{dvipsnames,svgnames,table}{xcolor}
\usepackage{hyperref}
\hypersetup{
  colorlinks=true,
  linkcolor=magenta,
  citecolor=magenta,
  urlcolor=blue!80!black,
  filecolor=blue!80!black,
}
\usepackage{natbib}
\usepackage{cleveref}

\usepackage{newfloat}
\usepackage{listings}
\DeclareCaptionStyle{ruled}{labelfont=normalfont,labelsep=colon,strut=off} % DO NOT CHANGE THIS
\floatstyle{ruled}
\newfloat{listing}{tb}{lst}{}
\floatname{listing}{Listing}
\newcommand{\paperN}{AoP-SAM\xspace}

\title{\paperN: Automation of Prompts for Efficient Segmentation}
\author {
    Yi Chen,
    Mu-Young Son,
    Chuanbo Hua,
    Joo-Young Kim
}
\affiliations {
    KAIST, Korea Advanced Institute of Science and Technology, Daejeon, 34141, South Korea \\
    \{chenyi, kkt1690, cbhua, jooyoung1203\}@kaist.ac.kr
}

\usepackage{bibentry}
\begin{document}

\maketitle

\begin{abstract}
The Segment Anything Model (SAM) is a powerful foundation model for image segmentation, showing robust zero-shot generalization through prompt engineering. 
However, relying on manual prompts is impractical for real-world applications, particularly in scenarios where rapid prompt provision and resource efficiency are crucial.
In this paper, we propose the Automation of Prompts for SAM (\paperN), a novel approach that learns to generate essential prompts in optimal locations automatically. \paperN enhances SAM’s efficiency and usability by eliminating manual input, making it better suited for real-world tasks. Our approach employs a lightweight yet efficient Prompt Predictor model that detects key entities across images and identifies the optimal regions for placing prompt candidates. This method leverages SAM’s image embeddings, preserving its zero-shot generalization capabilities without requiring fine-tuning. 
Additionally, we introduce a test-time instance-level Adaptive Sampling and Filtering mechanism that generates prompts in a coarse-to-fine manner. This notably enhances both prompt and mask generation efficiency by reducing computational overhead and minimizing redundant mask refinements.
Evaluations of three datasets demonstrate that \paperN substantially improves both prompt generation efficiency and mask generation accuracy, making SAM more effective for automated segmentation tasks.
% The Segment Anything Model (SAM) has proven to be a powerful foundation model for segmentation, demonstrating robust zero-shot generalization through prompt engineering. 
% However, relying on manual prompts is impractical for real-world applications, especially on edge devices that require rapid prompt provision and resource efficiency. 
% In this paper, we propose the Automation of Prompts for SAM (AoP-SAM), a novel approach that learns to predict the locations of essential prompts and is integrated with the Adaptive Sampling and Filtering (ASF) technique. 
% This method automatically generates essential prompts for segmentation, thereby eliminating the need for manual prompt provision. 
% We employ a simple yet efficient model that detects essential entities from entire images and identifies the optimal locations of potential prompts for segmentation, leveraging SAM’s image embeddings while ensuring its zero-shot generalization without fine-tuning. 
% Moreover, we propose ASF to collect essential prompts in a coarse-to-fine manner, significantly improving prompt and mask generation efficiency by greatly reducing redundant mask refinements. 
% We evaluate AoP-SAM on three datasets and demonstrate that our approach substantially enhances prompt and mask generation efficiency compared to previous prompting methods while matching the performance of segmentation. This improves the accuracy and efficiency of SAM in segmentation applications on edge devices.
\end{abstract}


\section{Introduction}
\label{sec:intro}

\begin{figure*}[t] 
    \centering
    \includegraphics[width=\linewidth]{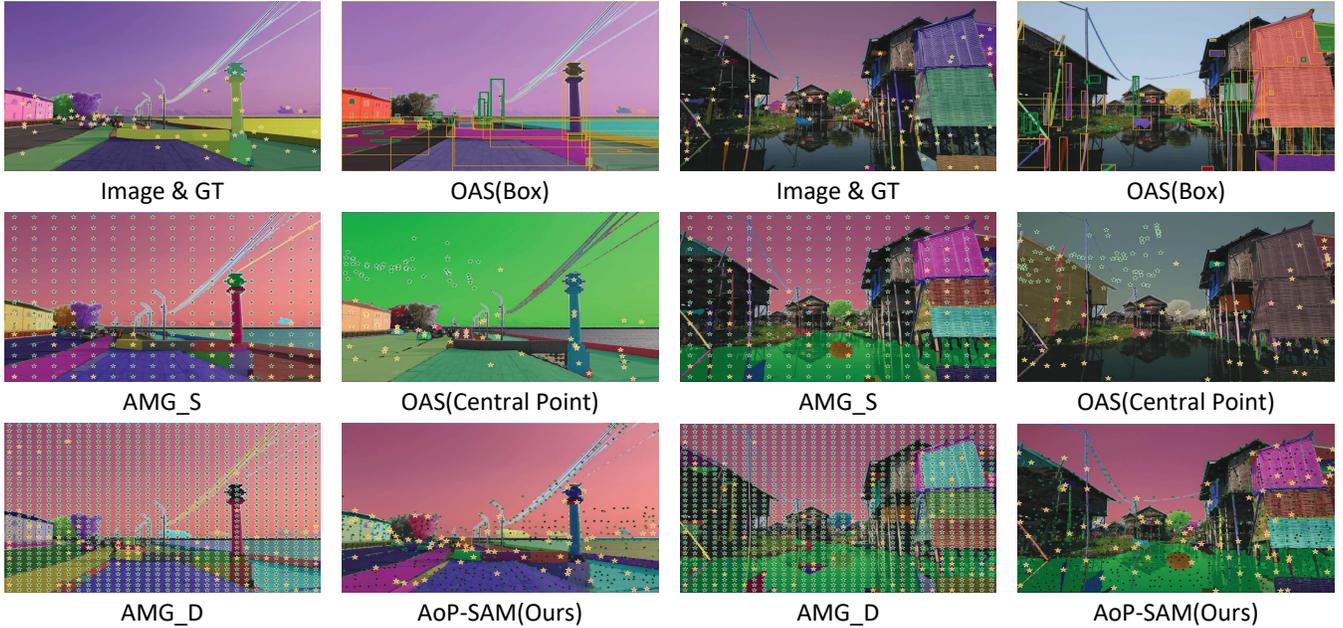}

    \caption{
    In SAM, automating prompt provision eliminates the need for manual input, significantly improving the efficiency of mask segmentation. However, current approaches, such as grid-based prompts in vanilla SAM (AMG\_S for sparse, AMG\_D for dense) or extra detection models (OAS: Box or Central Point), often introduce excessive mask refinements or computational overhead, leading to increased latency and reduced efficiency. In contrast, our proposed \paperN efficiently generates essential prompts for accurate mask generation within SAM, entirely without human intervention. In the illustrations above, different colors represent various segmentation results, with orange labels (stars or boxes) indicating valid prompts, green labels marking invalid prompts, and black stars in our results representing the filtered prompts, processed in a coarse-to-fine manner by the test-time instance-wise Adaptive Sampling and Filtering (ASF) mechanism.}
    \label{fig:Intro} 
\end{figure*}

Image segmentation, a critical task in CV, underpins applications ranging from autonomous vehicle navigation~\citep{feng2020deep} to medical diagnostics~\cite{hesamian2019deep} and robotics perception~\cite{hurtado2022semantic}.
Segment Anything Model (SAM) is a~\textit{foundation model} designed to tackle general image segmentation and has been trained on a vast dataset with billions of mask annotations~\cite{kirillov2023segany}. SAM excels at segmenting a wide range of visual elements across diverse environments, enabling it to solve various downstream segmentation problems through prompt engineering. These prompts, which include points or bounding boxes, allow SAM to achieve~\textit{zero-shot} generalization~\cite{kirillov2023segany}, making it adaptable to numerous applications.

The manual provision of prompts required for segmenting entire images in SAM is highly labor-intensive and time-consuming, making it impractical for applications that demand rapid prompt generation in hardware-constrained scenarios, such as industrial automation. Consequently, automatic prompt generation is essential for these use cases. However, as shown in Figure~\ref{fig:Intro}, current approaches to automating prompts for generalized tasks face significant limitations due to two key issues: 1) Unintelligent automation, the vanilla SAM employs a grid-based search of point prompts, named Automatic Mask Generation (AMG)~\cite{kirillov2023segany}, for producing prompts. If the grid search is too sparse, it risks missing numerous small objects or important details. Conversely, if the search is too dense, it produces an excessive number of redundant masks, necessitating significant refinement and ultimately slowing down the overall processing time. 2) Time and Resource Inefficiency, automating prompts with bounding boxes in SAM enables the use of existing deep-learning models to generate bounding boxes from images, offering an alternative for prompt generation. For instance, the Object-Aware Sampling (OAS) method from~\cite{zhang2023mobilesamv2} utilizes YOLOv8~\cite{wang2023uav}, a state-of-the-art architecture known for efficient detection with bounding boxes, to automate the prompt production process. However, this approach is not directly aligned with SAM and introduces substantial computational overhead, posing challenges in resource-limited scenarios. These constraints significantly diminish the applicability and effectiveness of foundational segmentation models like SAM, particularly in automated annotation tasks and situations where rapid prompt generation is crucial.

In this work, we propose a novel approach, \paperN, for the Automation of Prompts within the SAM family of models, e.g.~\cite{kirillov2023segany,mobile_sam,zhang2023mobilesamv2}. This method enables the efficient generation of essential point prompts for accurate segmentation without the need for human intervention.
We first designed a learnable prompt predictor specifically for SAM. Unlike independent deep learning modules that rely solely on image input to generate bounding boxes as prompt inputs, our predictor is tightly integrated with SAM. It takes both the image input and the computed image embedding—i.e., the input and output of SAM's image encoder—leveraging this information to learn and generate a prompt confidence map. This map predicts the locations of essential prompts that can be used for accurate segmentation within SAM. Second, to reduce the number of redundant mask generations, we propose an Adaptive Sampling and Filtering (ASF) technique that operates in a coarse-to-fine manner. Initially, we sample point prompts coarsely from the Prompt Confidence Map to begin mask generation. Then, we leverage the generated mask to filter out any remaining prompts that would produce the same mask, thereby enhancing overall efficiency. Therefore, \paperN can effectively and efficiently produce prompts for segmentation tasks without compromising the accuracy and flexibility of the original SAM. 

Our contributions are as follows:
\begin{itemize}
    \item To eliminate the need for manual provision of prompts tailored to each general image, our \paperN approach automatically predicts and generates prompts effectively and efficiently.
    \item We introduce a simple yet efficient prompt prediction method that utilizes SAM's computed data to pinpoint potential prompt locations. Additionally, ASF ensures that only the most essential prompts are utilized in a coarse-to-fine selection process, thereby enhancing overall segmentation efficiency.
    \item Extensive experiments on three benchmarks have shown the effectiveness of our proposed \paperN.
    
\end{itemize}

\section{Related Work}
\label{sec:rw}

% \subsection{Segment Anything Model}
\subsection{Prompting Technique in Zero-shot Foundation Models}
Foundation models initially emerged in NLP, with large language models like the GPT series demonstrating strong zero-shot generalization to unseen tasks and data. Prompt-based learning methods were then introduced, enabling these models to generalize to downstream tasks by interpreting prompts as task instructions rather than requiring parameter fine-tuning. The leading hypothesis regarding the effectiveness of prompts suggests that models interpret these prompts as specific task instructions, enabling them to generalize to tasks not encountered during training~\cite{sanh2021multitask}. This approach, inspired by human-like adaptability, quickly gained popularity in NLP~\cite{brown2020language}. These advancements influenced CV, where prompt engineering with frozen pre-trained models led to SAM, excelling in \textit{zero-shot} learning and precise object segmentation based on spatial prompts.

% These demonstrations are examples provided to the model to guide its understanding of a new task, and this learning ability is crucial as it underscores the models' flexibility in adapting to new information, mirroring human learning processes and soon prompting has been wildly used in NLP~\cite{brown2020language}.
% These advancements in NLP have extended their influence into CV, and prompt engineering that freezes the pre-trained model led to the creation of SAM. SAM showcases exceptional \textit{zero-shot} learning capabilities and ability to precisely segment objects based on spatial cues of prompts. 

\subsection{Methods for Automating Prompts}
While SAM allows manual prompts for mask generation (e.g., clicking or dragging on an image), this approach is impractical for real-world applications. The manual provisions of prompts required by SAM are highly labor-intensive and time-consuming. Moreover, the segmentation performance is heavily dependent on the prompt quality. Crafting precise prompting needs expert domain-specific knowledge, which is not available for all circumstances. To address this, SAM introduces an Automatic Mask Generation (AMG) mode, which autonomously positions numerous prompts in a grid-search manner and generates masks without continuous human input~\cite{kirillov2023segany}. However, sparse grids may miss small objects, while dense grids (e.g., $32 \times 32$ points) result in redundant prompts for large objects, requiring post-filtering. A special version of SAM trained for fully automatic mask generation created the extensive SA-1B dataset~\cite{kirillov2023segany} (11 million images, over 1 billion masks) but sacrifices inference speed, further increasing latency.

There is another direction currently by using modern object detection models to generate object-aware prompts~\cite{zhang2023mobilesamv2} adopts YOLOv8, which is a SOTA architecture for efficient detection with bounding boxes. With the generated box, people can either use its center as an object-aware point prompt or directly adopt the box itself as the prompt. However, this method brings heavy computational overhead due to the size of the object detection models, as they are not specialized for generating prompts. In contrast, our proposed approach is dedicated to predicting prompts and integrated ASF to further improve efficiency by relieving potential redundant generation through a sample-level test-time adaptation.

\subsection{Test-time Adaptation} Test-time domain adaptation aims to improve model performance on test data that differs from the training data due to a domain gap~\cite{wang2020tent,hu2020discriminative}. This adaptation is categorized into two main approaches: backward-based and backward-free. Backward-based adaptation utilizes self-supervised learning, often through entropy minimization, to learn the characteristics of the target domain~\cite{wang2020tent,hu2019multi}. In contrast, backward-free adaptation relies on batch normalization statistic adjustments, as demonstrated by DUA’s running average technique~\cite{mirza2022norm} and DIGA’s distribution adaptation for semantic segmentation~\cite{wang2023dynamically}. Previous research has also explored test-time domain adaptation for camouflage object segmentation in SAM using a general task description~\cite{hu2024relax}. Similarly, in our work, we implement instance-level test-time domain adaptation, focusing on adaptively removing redundant prompt candidates. This method enhances mask generation efficiency across diverse datasets without requiring sample-level supervision.

\section{Method}
\label{sec:method}

\begin{figure*}[t] 
    \centering
    \includegraphics[width=\linewidth]{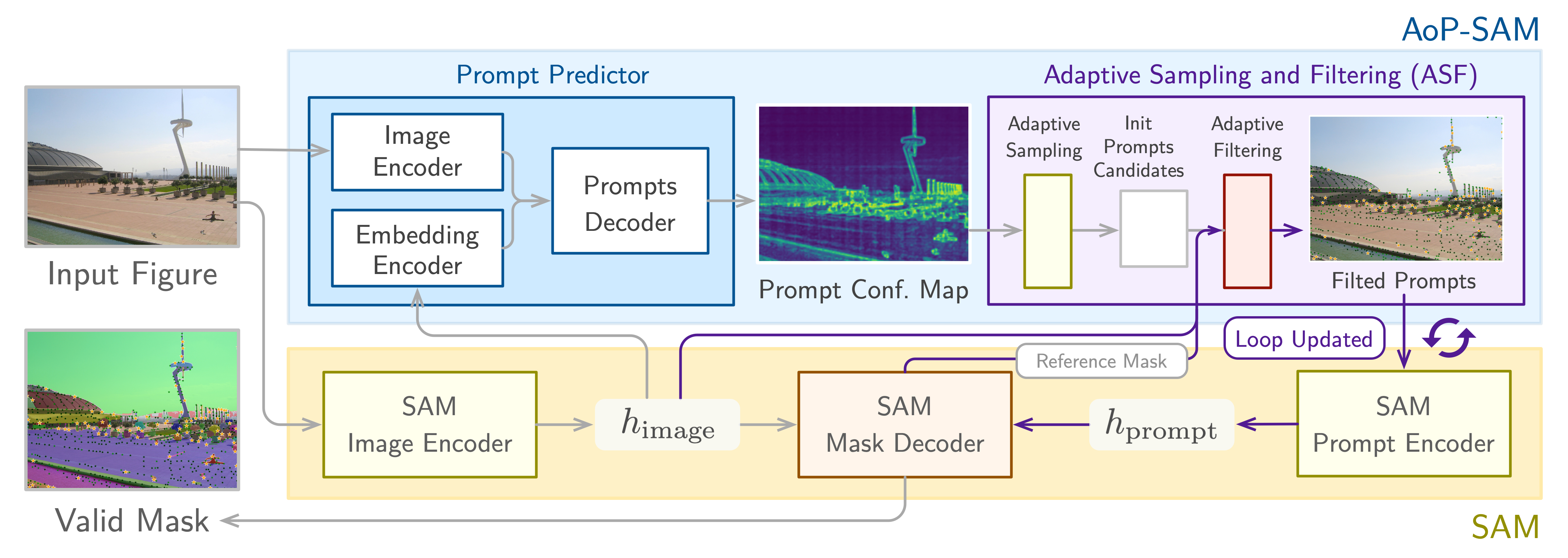}

    \caption{The architecture of our proposed \paperN consists of two key components: the prompt predictor and the Adaptive Sampler and Filter (ASF) Module. The prompt predictor operates by taking the image input and the computed image embedding from SAM's image encoder as inputs. Prompt predictor then generates a Prompt Confidence Map (PCM) that highlights potential regions for prompt candidates. During test-time, these candidates are adaptively sampled and filtered by ASF, predicting prompts that might lead to redundant masks based on the generated mask references. This process eliminates unnecessary prompts, ensuring that only the essential ones are used to generate the final mask results.}
    \label{fig:Overall} 
\end{figure*}

We propose \paperN to efficiently produce essential prompts for accurate mask generations in SAM. 
In this section, we first briefly review the architecture of SAM to show how our proposed Prompt Predictor collaborates with SAM. Then we introduce our Prompt Predictor, which identifies essential prompt locations that contribute to segmentation performance and further derives a prompt confidence map to guide prompt generation. We also describe the training and inference process of the Prompt Predictor. which is both data and computationally efficient. Lastly, we present the ASF technique to sample and filter prompts during the test-time adaptation.

\subsection{Preliminaries: SAM} SAM is an advanced image segmentation framework composed of three key components: an Image Encoder, a Prompt Encoder, and a Mask Decoder. These modules collaborate to process images and generate segmentation masks. (1) Image encoder: SAM begins with a robust yet computationally intensive module that extracts essential features from the input image, producing a 64×64 spatial resolution embedding as a compact representation of critical image characteristics. (2) Prompt encoder: The Prompt Encoder processes interactive inputs like points, boxes, or masks, converting them into embeddings that guide the Mask Decoder. This enhances accuracy and supports SAM’s remarkable zero-shot generalization. (3) Mask decoder: In the final stage, a two-layer transformer-based module that combines image and prompt embeddings to generate precise segmentation masks, effectively delineating objects or regions of interest. SAM optimizes efficiency by embedding image and prompt inputs only once. 

SAM’s zero-shot generalization is underpinned by the SA-1B dataset, containing over 1 billion masks and 11 million images—400 times larger than prior segmentation datasets. This extensive dataset allows SAM to segment new images without additional training. However, training SAM is resource-intensive: for instance, training the ViT-H-based SAM model on SA-1B for two epochs requires 256 GPUs and a batch size of 256 images~\cite{kirillov2023segany}, emphasizing the significant resources dedicated to the image encoder. This high computational cost motivates the reuse of the image encoder’s outputs in later computations to maximize efficiency. For more details, see ~\cite{kirillov2023segany}.

\subsection{Prompt Predictor for Prompt Confidence Map}
To enhance the automation of prompt production, we address two main challenges: the detection of essential entities within full images and the efficient identification of potential prompt locations. We propose a lightweight Prompt Predictor that integrates processed data from the Segment Anything Model (SAM), reducing computational complexity while maintaining tight coupling with SAM, which improves system efficiency. As illustrated in Figure~\ref{fig:Overall}, the image segmentation process begins with the computation of image embeddings. Whenever new prompts are provided, corresponding masks can be generated, meaning that prompt embeddings are computed and then injected into the mask decoder to produce these masks. Following recent methodologies, we initiate prompt generation after the image embedding is completed. This approach allows our Prompt Predictor to reuse pre-processed image inputs and their corresponding embeddings to generate a Prompt Confidence Map. This map identifies regions of high confidence, which can then be selected as prompts and used for segmentation within the SAM framework.

\textbf{Prompt Predictor Architecture.} The model employs two distinct CNN-based encoders: one for processing the original image and the other for handling the ViT embedding. The image encoder captures spatial features through three convolutional layers, each followed by ReLU activations to enhance feature extraction and introduce non-linearity. For the ViT embedding, the process begins by reshaping it to match the spatial dimensions of the image input, ensuring seamless integration with the image data during decoding. After reshaping, the ViT embedding is passed through its own encoder, transforming it into a 32-channel feature map. This alignment of the ViT embedding with the image's feature space allows the model to effectively fuse and decode information from both inputs, leveraging their strengths for more accurate and robust outcomes.

Following the encoding process, the Prompt Predictor concatenates the outputs from both the image encoder and the ViT embedding encoder along the channel dimension. This operation merges the feature maps, creating a fused representation that integrates both spatial and contextual information from the original image and the ViT embedding. The fused feature map is then processed through a series of convolutional layers in the Prompt decoder. These layers progressively reduce the dimensionality of the combined features, refining the representation and distilling it into a more compact form that retains the most relevant information. The decoding process culminates in a Sigmoid-activated layer, producing the final output as a Prompt Confidence Map. This map effectively highlights regions of interest, from which prompts can be sampled. The Sigmoid function normalizes the output to a range between 0 and 1, making it well-suited for generating probability maps that allow for the flexible selection of high-confidence regions, ensuring the prompts generated are both relevant and useful for mask segmentation within the SAM framework.

The Prompt Predictor is designed with efficiency in mind, maintaining a small memory footprint and low computational burden. The use of lightweight CNN-based encoders for both the original image and the ViT embedding ensures efficient feature extraction without the need for overly complex architectures. Each encoder consists of a limited number of convolutional layers, reducing computational load while capturing essential features. By reshaping the ViT embedding to match the image dimensions and aligning it within the same feature space, the model streamlines the process, avoiding unnecessary intermediate computations. The final step of concatenating outputs along the channel dimension, followed by a compact decoder, further minimizes resource usage. The decoder efficiently refines the fused feature map and produces the final Prompt Confidence Map through a single Sigmoid layer, which is computationally inexpensive. This efficient design allows the Prompt Predictor to deliver high performance while keeping memory usage and computational demands to a minimum, making it an ideal solution for automating prompt generation in a manner that is both resource-efficient and fast.

\textbf{Training of Prompt Predictor} Unlike traditional object detection models that typically train on datasets like COCO~\cite{lin2014microsoft}, our approach to training \paperN is both data-efficient and closely aligned with SAM by leveraging the SA-1B dataset~\cite{kirillov2023segany}. The SA-1B dataset, containing over 1 billion masks and their corresponding prompts, was specifically chosen to ensure that \paperN inherits the robustness and generalizability of SAM. By training on the same dataset, we align \paperN with SAM’s capabilities, particularly in handling diverse and unseen data. Given that point-type prompts align most effectively with the Prompt Confidence Map, which can be generated by our Prompt Predictor, we use these point prompts from the SA-1B dataset as ground truth to train our model. To ensure a diverse and challenging training set, we carefully curated a selection of samples from the SA-1B dataset, encompassing a broad range of semantic classes and complex scenarios. This careful curation is crucial for maintaining the model's ability to generalize effectively across various object scales and complexities.

During training, the learnable parameters include the Image Encoder, ViT Encoder, and Prompt Decoder. To generate the Prompt Location Map Ground Truth, we place point prompts within a blank map that matches the dimensions of the pre-processed image sample. Since the initial ground truth map is too sparse for effective training, we refine it using a combination of uniform and Gaussian kernels. This refinement enhances the precision of prompt locations, making the training feasible and significantly improving the prompt generation process.  The training process involved iterative refinement of the model's parameters using a MSELoss function, with optimization carried out via the Adam optimizer. We employed a learning rate of and trained the model for 1000 epochs, using gradient accumulation to handle larger batch sizes effectively. 

\subsection{ASF for Essential Prompts}
To effectively sample prompt candidates from the Prompt Confidence Map, a Gaussian filter is first applied to smooth the confidence map, enhancing key regions while reducing noise. Following this, local maxima within the smoothed confidence map are identified by isolating critical points of interest, taking into account both a minimum distance between peaks and an absolute threshold. These local maxima represent potential prompt candidates. The identified points, initially located within the resized output dimensions, are then mapped back to the original image coordinates by scaling them according to the ratio between the resized output and the original image size. This mapping results in a set of coordinates that accurately reflect the positions of significant features within the original input space, making them suitable as prompt candidates.

Even though sampling the candidates from the Prompt Confidence Map can make a good selection for Automating Prompts, however, it is still possible to make some redundant prompts in some cases, therefore, we adapt a further fine filtering to remove redundant candidates. In the processes of mask generation in SAM, due to memory constraints, the prompt can be divided into several batches and processed iteratively, therefore, we can take advantage of computated masks as references to predict which prompts remained in the prompt candidates will be redundant and result in the same mask with them, so we need to figure out the spatial location and semantic meaning of these generated masks first and then obtain the Prompt Elimination Map~\cite{zhang2023personalize} of the processed prompts.  

We utilize the image feature map result from the vit\_embedding and also reference masks during mask generation, where the image feature map containing the original image input information and mask data containing the information of location are generated, we denote image feature map as $F_{\text{feat}}\in \mathbb{R}^{h \times w 
 \times c}$ and each of these $n$ masks as $M_{\text{ref}}\in \mathbb{R}^{h \times w }$ with $h$, $w$ denoting the dimension of the image feature map, $c$ as the feature dimension and $n$ is the number of predicted masks. 
 
 The $n$ down-sampled reference masks $M_{\text{ref}}$ are used to extract the mask feature $M_{\text{feat}}$ from the image feature map one by one and we can get a set of $n$ mask features. Each mask feature then adopts an average pooling to aggregate its global visual embedding. After this, we can obtain a Prompt Elimination Map with confidence $C$  for each reference mask by doing a cosine similarity between pixel-wisely L2-normalized  mask and image feature $M_{\text{norm}}$ and $F_{\text{norm}}$ as
\begin{equation}
\{C^i\}_{i=1}^{n} = \{F_{\text{norm}} \times M_{\text{norm}}^i\}_{i=1}^{n}\text{ , }  N_{\text{norm}} \in \mathbb{R}^{ n \times h \times w}
\end{equation}

On top of this, we adopt another average pooling to aggregate all $n$ local maps to obtain the overall Prompt Elimination Map of the generated mask as 

\begin{equation}
C = \frac{1}{n} \sum_{i=1}^{n} C_i \text{ , }C \in \mathbb{R}^{ h \times w}
\end{equation}

\begin{table*}[ht]
\centering
{\fontsize{10}{12}\selectfont
\adjustbox{max width=\textwidth}{
\begin{tabular}{l|l|cccc|cccc|cccc}
\toprule
% \multirow{2}{*}{\centering Image Encoders} & \multirow{2}{*}{Automating Prompts Methods} & \multicolumn{4}{c|}{SA-1B} & \multicolumn{4}{c|}{COCO} & \multicolumn{4}{c}{LVIS} \\
\multirow{2}{*}{\centering Image Encoders} & \multirow{2}{*}{Auto Prompts Methods} & \multicolumn{4}{c|}{SA-1B} & \multicolumn{4}{c|}{COCO} & \multicolumn{4}{c}{LVIS} \\
\cmidrule(lr){3-6} \cmidrule(lr){7-10} \cmidrule(lr){11-14}
 &  & $\text{mIoU} \uparrow$ & $\text{Inf}_{\text{Lat.}} \downarrow$ & $\text{Peak}_{\text{Mem.}} \downarrow$& $\#{P}$  & $\text{mIoU} \uparrow$ & $\text{Inf}_{\text{Lat.}} \downarrow$ & $\text{Peak}_{\text{Mem.}} \downarrow$ & $\#{P}$ & $\text{mIoU} \uparrow$  & $\text{Inf}_{\text{Lat.}} \downarrow$ & $\text{Peak}_{\text{Mem.}} \downarrow$ & $\#{P}$  \\
\midrule
\multirow{5}{*}{\centering MobileSAM} 
& AMG\_S~ & 29.8 & - & 4.5 & 38.6 &  56.0 & - & \textbf{1.9} & 33.5 & 56.2 & -  & \textbf{1.9} & 33.4  \\ 
& AMG\_D~ & 46.9 & - & 9.1 & 71.0 &  60.9 & - & \textbf{1.9} & 55.9 & 61.1 & -  & \textbf{1.9} & 55.5   \\ 
& OAS(Box)~ &  50.7&  0.191&  7.3& 100& 55.5 & 0.187& 4.2& 44&  55.7& 0.188&  4.0 & 38\\
& OAS(Central Point)~ & 48.7 &  0.188& 7.7 & 141.0 &  53.9&  0.167&  4.3&  69.0&  54.5&  0.164&  4.3 &  68.1\\
& AoP-SAM & \textbf{51.4} &  \textbf{0.101}& \textbf{4.1} &  71.7 & \textbf{61.5} &  \textbf{0.096}& 2.1 & 58.1 & \textbf{62.3} & \textbf{0.094}& 2.1 & 57.5 \\ 

\midrule

\multirow{5}{*}{\centering ViT\_L} 
& AMG\_S~ &  40.0 & - & 5.7 & 55.5 &  61.4 & - & 4.4 & 48.8 & 63.2 & - & \textbf{4.3} & 49.5 \\
& AMG\_D~ & 65.6 & - & 10.3 & 108.9 &  67.7 & - & \textbf{4.3} & 86.0  & 69.2 & - & \textbf{4.3} & 86.5  \\
& OAS(Box)~ &  65.8 &  0.150&  9.1 &  100&  63.3&  0.152&  5.4 & 44&  62.9&  0.151&  5.3 &  38\\
& OAS(Central Point)~ &  67.6&  0.149&  9.7& 199.3&  64.2&  0.133&  5.5&  98.4&  63.5&  0.132& 5.5& 98.9\\
& AoP-SAM & \textbf{71.1} & \textbf{0.120}& \textbf{5.4} & 118.3  & \textbf{68.4} &  \textbf{0.116}& 4.4 & 97.0 & \textbf{69.8} & \textbf{0.117} & 4.4 & 97.2\\
\midrule

\multirow{5}{*}{\centering ViT\_H} 
& AMG\_S~ & 40.8 & - & 7.1& 56.3 &  63.3 & - & 5.7& 49.8 & 64.9 & - & 5.6 & 50.5\\
& AMG\_D~ & 66.8 & - & 11.8& 109.6 & 69.5 & - & 5.7 & 87.4 & 71.0 & - & 5.6 & 88.0 \\
& OAS(Box)~ &  66.9 & 0.160& 10.4 & 100 & 64.1& 0.152&  6.8 & 44& 63.3&  0.153&  6.6 & 38\\
& OAS(Central Point)~ &  68.3&  0.154&  11.1 & 207.6 &  65.1 & 0.134&  6.9& 102.1& 63.0&  0.134& 6.8 & 102.4\\
& AoP-SAM & \textbf{70.6} &  \textbf{0.122}&  \textbf{6.6} & 107.8 & \textbf{70.1} &  \textbf{0.120}&  \textbf{5.5} &  90.0 & \textbf{71.9} & \textbf{0.122}& \textbf{5.5} & 89.7 \\

\bottomrule
\end{tabular}
} % Closing the \adjustbox
}
\normalsize 
\caption{Results on Image Segmentation with bounding box supervision and point supervision. Best are in \textbf{bold}.}
\label{tab:results}
\end{table*}

By incorporating the Prompt Elimination Maps of every high-quality mask, the Elimination Map can take the visual appearance of different objects from existing masks into consideration, and acquire a relatively comprehensive location estimation. Each pixel on the upsampled Elimination Map has a Elimination Score that indicates the likelihood of that pixel is in the same spatial location and having the same semantic meaning as generated masks in the original image, The higher the Elimination Score of each pixel is in the map, the more likely prompting at that pixel will resulting in the duplicated masks with the existing masks.

After getting the Elimination Map generated from all existing masks, we can then obtain the elimination threshold $T_{\text{elim}}$ by calculating the Elimination Score of current processing prompts along with the confidence Intersection over Union (IoU) scores of their resulting masks, as follows:

\begin{equation}
T_{{\text{elim}}} = \frac{1}{n} \sum_{i=1}^{n} {{\text{IoU}}^{i} \times C^{i} } 
\end{equation}

By multiplying the IoU score of the masks generated by prompts with the Elimination Scores of current processing prompts, the threshold biases more with prompts which can generate the higher-quality masks.

The remaining prompts in the pool will obtain their corresponding Elimination Scores, which are compared to the threshold. If a prompt’s score exceeds the threshold, it is considered redundant and would generate duplicate masks if further processed; therefore, it should be eliminated from the prompt pool. By doing this way, redundant prompts will be eliminated and only essential prompts will be kept in the prompt pool and used to generate masks in the following iterations.

\section{Experiments}
\label{sec:exp}

\begin{table*}[ht]
\centering
{\fontsize{10}{12}\selectfont
\adjustbox{max width=\textwidth}{
\begin{tabular}{cccccccccccccccc}
\toprule
\multicolumn{3}{c}{\textbf{Method's variant}} & \multicolumn{4}{c}{SA-1B} & \multicolumn{4}{c}{COCO} & \multicolumn{4}{c}{LVIS} \\
\cmidrule(lr){1-3} \cmidrule(lr){4-7} \cmidrule(lr){8-11} \cmidrule(lr){12-15}
  Prompt Predictor & Adaptive Sampling  & Adaptive Filtering & $\text{mIoU} \uparrow$ & $\text{Inf}_{\text{Lat.}} \downarrow$ & $\text{Peak}_{\text{Mem.}} \downarrow$ & $\#{P}$ & $\text{mIoU} \uparrow$ & $\text{Inf}_{\text{Lat.}} \downarrow$ & $\text{Peak}_{\text{Mem.}} \downarrow$ & $\#{P}$ & $\text{mIoU} \uparrow$ & $\text{Inf}_{\text{Lat.}} \downarrow$ & $\text{Peak}_{\text{Mem.}} \downarrow$ & $\#{P}$
\\
\midrule
 $\checkmark$  &  &   & 57.2 & 0.059 & 7.2 & 106.4 & 67.9 & 0.078 & 5.7 & 70.4 & 60.9  & 0.075 & 5.7 & 60.6 \\
 $\checkmark$ & $\checkmark$ &   & 72.8 & 0.130 & 10.1 & 120.1 & 70.5 & 0.122 & 5.7  & 97.9 & 71.7 & 0.121 & 5.7 & 97.5 \\
 % $\checkmark$ & $\checkmark$ & $\checkmark$ &  & 0.085 & 0.699 & 0.807 & 0.754 & 0.143 & 0.537 & 0.746 & 0.666 & 0.106 & 0.681 & 0.838  \\
 $\checkmark$ & $\checkmark$ & $\checkmark$ & 71.3 &  0.122 &  6.6 & 107.8 & 70.1 & 0.112&  5.7 &  91.1& 71.9& 0.122& 5.7 & 89.7\\
\bottomrule
\end{tabular}
}
}
\caption{Ablation study of variants with our \paperN on image segmentation.}
\label{tab:ablation}
\end{table*}

To evaluate \paperN across various scenarios, we selected three different image encoders and implemented five automating prompts methods. This approach allows us to comprehensively assess both the accuracy and efficiency of our method under different conditions.

\subsection{Setup}
\textbf{Datasets.} Generalized image segmentation focuses on segmenting every meaningful entity in an image. In this study, we use three key datasets: SA-1B, COCO, and LVIS. The SA-1B dataset, used for training SAM, contains over 1 million images and 1 billion masks~\cite{kirillov2023segany}. The COCO dataset includes 41,000 images and 200,000 masks, covering a wide range of common objects~\cite{lin2014microsoft}. LVIS, designed for long-tail distributions, provides 5,000 images and 25,000 masks, emphasizing fine-grained categories~\cite{gupta2019lvis}. These datasets allow us to thoroughly evaluate the effectiveness of our Automating Prompts method across diverse and challenging scenarios.

\textbf{Baseline.} In our comparison of current methods for Automating Prompts in SAM, we introduce and evaluate two types of prompts: bounding box prompts and point prompts. The methods AMG\_S and AMG\_D represent the vanilla grid search with $16\times16$ and $32\times32$ prompts, respectively, as utilized in SAM~\cite{kirillov2023segany}. We also examine the Object-Aware Sampling (OAS) method, which employs YOLOv8 to generate bounding box prompts~\cite{zhang2023mobilesamv2}. Furthermore, we implement an additional method that uses the central point of the bounding box generated by OAS as point prompts. Note that \paperN is trained on a subset dataset of SA\_1B and tested on a separate test set, similarly all the comparative methods we employ are also trained and tested on different sets. 

Evaluating SAM’s accuracy is challenging as it generates masks without predefined labels, making traditional metrics like $\text{mIoU}$~\cite{shotton2006textonboost,han2023segment}, $mAP$~\cite{lin2014microsoft,henderson2017end}, and $PQ$\cite{kirillov2019panoptic} unsuitable~\cite{zhang2023survey}. To address this, we use the greedy IoU algorithm~\cite{zhang2023survey}, which matches each SAM mask with the closest ground truth mask based on IoU and calculates the mean IoU ($\text{mIoU}$) for all matches. In addition to evaluating accuracy performance, we also assess the efficiency of Methods of Automating Prompts in time- or resource-constrained environments using Inference Latency ($\text{Inf}_{\text{Lat.}}$)(s) for producing prompts and peak memory ($\text{Peak}_{\text{Mem.}}$)(GB) consumption during mask generation as key metrics. Additionally, we count the number of essential prompts ($\#P$) as a reference point for comparing methods. It is important to note that a higher value of $\text{mIoU}$, or lower values of $\text{Inf}_{\text{Lat.}}$ and $\text{Peak}_{\text{Mem.}}$, indicate higher efficiency. Although there is no clear preference for the number of essential prompts, intuitively, a smaller number of prompts yielding high accuracy performance is considered advantageous.

\begin{table*}[h!]
\centering
{\fontsize{10}{12}\selectfont
\adjustbox{max width=\textwidth}{
\begin{tabular}{cccc}

    % (a) Sampling Smoothing Factor
        \begin{tabular}{c|c|c|c}
            \multicolumn{4}{c}{(a) Sampling Smoothing Factor } \\
            \toprule
            Factor & $\text{mIoU} \uparrow$ & $\text{Inf}_{\text{Lat.}} \downarrow$ & $\text{Peak}_{\text{Mem.}} \downarrow$ \\
            \midrule
            1 & \textbf{72.4}  & 0.124 & 51.5  \\
            \cellcolor{gray!40}2 & \cellcolor{gray!40}70.4  & \cellcolor{gray!40}0.122 & \cellcolor{gray!40}42.2\\
            3 & 67.3  & \textbf{0.118} & 32.7 \\
            4 & 63.4 & 0.122 & \textbf{24.6}\\
            % 5 & 0.066 & 0.683 & 0.833 & 0.775 \\
            \bottomrule
        \end{tabular}
    % (b) Prompt Peak Spacing 
        \begin{tabular}{c|c|c|c}
            \multicolumn{4}{c}{(b) Confidence Intensity Threshold} \\
            \toprule
           Thr. & $\text{mIoU} \uparrow$ & $\text{Inf}_{\text{Lat.}} \downarrow$ & $\text{Peak}_{\text{Mem.}} \downarrow$ \\
            \midrule
            0.1 & \textbf{70.9}  & 0.121 & 10.1 \\
            \cellcolor{gray!40}0.2 & \cellcolor{gray!40}70.4  & \cellcolor{gray!40}0.122 & \cellcolor{gray!40}9.75 \\
            0.3 & 68.7 & \textbf{0.116} & \textbf{9.52}\\
            0.4 & 66.4 & 0.117 & 9.60 \\
            % 0.5 & 0.107 & 0.595 & 0.753 & 0.708 \\
            \bottomrule
        \end{tabular}
    % (c) Heatmap threshold
        \begin{tabular}{c|c|c|c}
            \multicolumn{4}{c}{(c) Prompt Spacing Factor} \\
            \toprule
            Factor & $\text{mIoU} \uparrow$ & $\text{Inf}_{\text{Lat.}} \downarrow$ & $\text{Peak}_{\text{Mem.}} \downarrow$ \\
            \midrule
            4 & \textbf{72.7} & 0.123 & 10.0 \\
            5 & 71.6 & 0.123 & 9.88 \\
            \cellcolor{gray!40}6 & \cellcolor{gray!40}70.4 & \cellcolor{gray!40}0.122 & \cellcolor{gray!40}\textbf{9.75} \\
            7 & 68.9 & \textbf{0.117} & 9.82  \\
            \bottomrule
        \end{tabular}
        \begin{tabular}{c|c|c|c}
            \multicolumn{4}{c}{(d) Prompt Elimination Threshold } \\
            \toprule
            Thr. & $\text{mIoU} \uparrow$ & $\text{Mask}_{\text{Lat.}} \downarrow$ & $\text{Ratio}_{\text{elim.}} \uparrow$\\
            \midrule
            1.25 & 68.4& \textbf{0.671} & \textbf{51.5} \\
            \cellcolor{gray!40}1.3 & \cellcolor{gray!40}70.4 & \cellcolor{gray!40}0.799 & \cellcolor{gray!40}42.3 \\
            1.35 & 71.6 & 0.93 & 32.7  \\
            1.4 & \textbf{72.2} & 1.04 & 24.6  \\
            % 5 & 0.066 & 0.683 & 0.833 & 0.775 \\
            \bottomrule
        \end{tabular}

\end{tabular}
} 
}
\normalsize
\caption{Ablation study on Hyper-parameters employed in \paperN. Best are in \textbf{bold}}
\label{tab:Hyper}
\end{table*}

\textbf{Implementation Details.}
Following the previous prompting settings~\cite{kirillov2023segany}, we enable the option for generating multiple mask outputs from a single prompt for point prompts, while disabling it for box prompts~\cite{zhang2023mobilesamv2}. No background prompts are provided in either case. We also implemented quality checks for all methods, removing low-quality masks (e.g., those with low confidence or stability scores) during performance evaluation. 

For coarsely sampling point prompts from the Prompt Confidence Map, we first apply a Smoothing Factor=2, a Confidence Intensity Threshold=0.2, and a Prompt Spacing Factor=2 as initialized parameters. In each iteration, the output mask from the previous iteration serves as a reference to generate a Prompt Elimination Map via the ASF, adaptively filtering out selected prompt candidates during test-time to prevent redundant mask generation in future iterations. The experiments are conducted using the PyTorch framework on a single Nvidia Titan RTX GPU.

\subsection{Experiment Results and Analysis}
\textbf{Experiment Results.} 
Table~\ref{tab:results} compares the performance of various Automating Prompt methods across different image encoders and evaluated on three datasets. Across all datasets and image encoders, \paperN consistently achieves the highest mIoU scores, even though bounding box methods inherently benefit from more spatial information. This underscores the effectiveness of \paperN in leveraging prompts for accurate segmentation, surpassing both traditional methods and those that rely on advanced object detection models. The \paperN method not only improves accuracy but also demonstrates competitive latency and memory usage. For example, on the SA-1B dataset with the ViT\_H encoder, \paperN achieves a latency of 0.122s and peak memory usage of 6.6GB, which are within acceptable ranges while delivering superior segmentation performance. Overall, the OAS methods (using either box or central point prompts) generally perform better than the baseline AMG\_S and AMG\_D methods but fall short of \paperN. This indicates that while object-aware sampling improves prompt effectiveness, the adaptive sampling and filtering techniques employed in \paperN further enhance the accuracy of segmentation and efficiency of Automating Prompts.

\textbf{Component Analysis.} We further analyze the impact of components including the Prompt Predictor, Adaptive Sampling, and Adaptive Filtering in Table~\ref{tab:ablation} on the same datasets. When Adaptive Sampling is enabled, there is a notable improvement in mIoU compared to using only the Prompt Predictor. However, the best performance is observed when both Adaptive Sampling and Adaptive Filtering are used together, highlighting the importance of filtering redundant prompts to enhance segmentation accuracy. The study shows that while the full \paperN configuration achieves the highest mIoU, it slightly increases latency and memory usage, a key trade-off for speed-sensitive applications.

\textbf{Sampling Smoothing Factor.} In Table~\ref{tab:Hyper}a, we apply Gaussian filtering to the heatmap using the Sampling Smoothing Factor. A larger Sampling Smoothing Factor allows the model to cover a broader area, providing more substantial smoothing, which is useful for reducing memory access during preparation and processing.

\begin{table}[ht]
\centering
\setlength{\tabcolsep}{2pt}
\adjustbox{max width=\columnwidth}{ 
{\fontsize{10}{12}\selectfont
\begin{tabular}{l|c|c|c|c}
\toprule
\multirow{2}{*}{\textbf{Prompt Methods}} & Latency (Sec) $\downarrow$ & Latency (Sec) $\downarrow$  & Latency (Sec) $\downarrow$ & Peak Mem $\downarrow$ \\
& (SA-1B) & (COCO) & (LVIS) & (GB) \\
\midrule
OAS(Box) & 1.16 & 1.01 & 0.99 & 0.78 \\
OAS(Central) & 1.32 & 1.21 & 1.23 & 0.78 \\
AoP-SAM & \textbf{0.65} & \textbf{0.77} & \textbf{0.84} & \textbf{0.042} \\
\bottomrule
\end{tabular}
}
}
\caption{Experimental results with MobileSAM for prompt automation efficiency on an Nvidia Jetson Orin Nano Edge GPU.}
\label{tab:edge}
\end{table}

\textbf{From heatmap to point prompt.} In Table~\ref{tab:Hyper}b-\ref{tab:Hyper}c, we explore various parameter settings to transform the confidence map into optimized point prompts. By adjusting the Confidence Intensity Threshold and Prompt Spacing Factor, we aim to identify the optimal points that most accurately represent the critical areas in the confidence map. These adjustments help refine the sensitivity of the point selection process, ensuring that the resulting point prompts are both precise and reliable.

\textbf{Prompt Elimination Threshold.} We evaluate the impact of the Prompt Elimination Threshold on the prompt removal ratio in Table~\ref{tab:Hyper}d. As the Prompt Elimination Threshold decreases, the prompt removal ratio increases, resulting in a speed-up effect of the mask generation while may slightly affect accuracy.

\textbf{Edge Device.} We conducted the prompt automation experiment on an Nvidia Jetson Orin Nano Edge GPU, obtaining the results in Table~\ref{tab:edge}. Due to hardware limitations, only MobileSAM~\cite{zhang2023faster} could run, as other pre-trained models exhausted the edge GPU memory. We focused on evaluating the efficiency of prompt automation, with accuracy expected to align with standard GPU results. These results further demonstrate \paperN's reliability on edge devices, achieving lower inference latency and reduced peak memory usage, making \paperN well-suited for deployment in resource-constrained environments.

\section{Conclusion}
\label{sec:Con}
We propose \paperN, a novel approach designed to efficiently generate essential prompts for accurate mask generation in SAM. Our method introduces a lightweight prompt predictor, which is trained to predict optimal prompt locations, complemented by a test-time adaptive sampling and filtering technique that automatically produces these prompts for SAM. We evaluate the accuracy and efficiency of \paperN on three segmentation datasets with three SAM family models. The results demonstrate that \paperN enhances both the accuracy and efficiency of SAM in generalized image segmentation tasks, making it ideal for automated prompt-based segmentation tasks with SAM.

% We propose \paperN, a novel approach designed to efficiently generate essential prompts for accurate mask generation in SAM. Our method features a lightweight Prompt Predictor, trained to predict optimal prompt locations, and a test-time ASF mechanism for automatic prompt generation. Evaluated on three segmentation datasets with three SAM family models, \paperN improves both accuracy and efficiency, making it ideal for automated prompt-based segmentation tasks with SAM.

\section{Acknowledgments}
\label{sec:acknowledgments}
This work was supported by Institute of Information \& Communications Technology Planning \& Evaluation (IITP) grant funded by the Korea governments(MSIT)(No. 2022-0-01036, Development of Ultra-Performance PIM Processor Soc with PFLOPS-Performance and GByte-Memory \& No.2022-0-01037, Development of High Performance Processing-In-Memory Technology based on DRAM) and the authors would like to express their sincere gratitude to Adiwena Putra for comments on an earlier draft.

\bibliography{aaai25}

% \input{1_Introduction}
% \input{2_Related_Work}
% \input{3_Method}
% \input{4_Experiments}
% \input{5_Conclusion}

% \subsection{Tables}

% Tables should be presented in 10 point roman type. If necessary, they may be altered to 9 point type. You must not use \texttt{\textbackslash resizebox} or other commands that resize the entire table to make it smaller, because you can't control the final font size this way.
% If your table is too large you can use \texttt{\textbackslash setlength\{\textbackslash tabcolsep\}\{1mm\}} to compress the columns a bit or you can adapt the content (e.g.: reduce the decimal precision when presenting numbers, use shortened column titles, make some column duble-line to get it narrower).

% Tables that do not fit in a single column must be placed across double columns. If your table won't fit within the margins even when spanning both columns and using the above techniques, you must split it in two separate tables.

\end{document}